\documentclass[conference]{IEEEtran}
\IEEEoverridecommandlockouts
\usepackage{caption}
\usepackage{subcaption}
\ifCLASSINFOpdf
\else
   \usepackage[dvips]{graphicx}
\fi
\usepackage{url}
\usepackage{amsmath}
\usepackage{amsthm}
\usepackage{amssymb}
\usepackage[ruled,vlined]{algorithm2e}
\usepackage{breqn}
\newtheorem{theorem}{Theorem}

\def\BibTeX{{\rm B\kern-.05em{\sc i\kern-.025em b}\kern-.08em T\kern-.1667em\lower.7ex\hbox{E}\kern-.125emX}}

\usepackage{graphicx,color}

\DeclareMathOperator*{\argmin}{arg\,min}
\usepackage{textcomp}
\makeatletter
\newcommand{\distas}[1]{\mathbin{\overset{#1}{\kern\z@\sim}}}%
\newsavebox{\mybox}\newsavebox{\mysim}
\newcommand{\distras}[1]{%
  \savebox{\mybox}{\hbox{\kern3pt$\scriptstyle#1$\kern3pt}}%
  \savebox{\mysim}{\hbox{$\sim$}}%
  \mathbin{\overset{#1}{\kern\z@\resizebox{\wd\mybox}{\ht\mysim}{$\sim$}}}%
}

\usepackage{cite}

\usepackage{subcaption}

\title{Smoothing ADMM for Non-convex and Non-smooth Hierarchical Federated Learning}
\author{Reza Mirzaeifard, Stefan Werner \thanks{This work was supported by the Research Council of Norway.}\\

Department of Electronic Systems, Norwegian University of Science and Technology, Norway
  \\E-mails: \{reza.mirzaeifard,  stefan.werner\}@ntnu.no}

\begin{document}
\maketitle
\begin{abstract}
This paper presents a hierarchical federated learning (FL) framework that extends the alternating direction method of multipliers (ADMM) with smoothing techniques, tailored for non-convex and non-smooth objectives. Unlike traditional hierarchical FL methods, our approach supports asynchronous updates and multiple updates per iteration, enhancing adaptability to heterogeneous data and system settings. Additionally, we introduce a flexible mechanism to leverage diverse regularization functions at each layer, allowing customization to the specific prior information within each cluster and accommodating (possibly) non-smooth penalty objectives. Depending on the learning goal, the framework supports both consensus and personalization: the total variation norm can be used to enforce consensus across layers, while non-convex penalties such as minimax concave penalty (MCP) or smoothly clipped absolute deviation (SCAD) enable personalized learning. Experimental results demonstrate the superior convergence rates and accuracy of our method compared to conventional approaches, underscoring its robustness and versatility for a wide range of FL scenarios.
\end{abstract}
\section{Introduction}
\label{sec:intro}
The rapid expansion of cyber-physical systems within the internet of things (IoT) ecosystem has necessitated a shift from centralized data processing to federated frameworks. This transition is driven by the significant computational demands, bandwidth constraints, and privacy concerns associated with traditional centralized methods \cite{chen2019quantile,zhou2022admm}. Federated learning (FL) has emerged as a pivotal solution, enabling collaborative model training across distributed devices while keeping data localized, thereby addressing critical privacy challenges and communication overhead \cite{yin2021comprehensive,zhou2023decentralized}. Recent advancements have further enhanced FL’s scalability through hierarchical architectures, where two layers of aggregation are performed: edge servers handle intra-set computations for individual client groups, while a cloud server manages inter-set coordination across clusters \cite{chen2020asynchronous,zhou2023decentralized}. This hierarchical approach optimizes resource utilization and facilitates parallel computations, making it indispensable for large-scale distributed systems.

Despite these benefits, hierarchical FL environments face substantial heterogeneity among participating clients, necessitating flexible update strategies. First, data heterogeneity arises from distribution skew, label skew, feature skew, and variations in data quality (e.g., quantity skew and noise levels) \cite{Kairouz2021Advances,Zhang2022Federated}. Second, model heterogeneity occurs when clients must tailor model architectures or optimizers to accommodate various hardware or domain requirements \cite{Diao2020HeteroFL,Li2019FedMD}. Third, task heterogeneity emerges when clients solve distinct yet related problems \cite{Smith2017Federated,Marfoq2021Federated}. Fourth, communication heterogeneity stems from inconsistent network conditions and bandwidth constraints \cite{Nishio2019Client,Lai2021Oort}. Finally, device heterogeneity introduces disparities in computational power and storage capacities \cite{shin2022fedbalancer,li2020federated}. These factors underscore the need for a flexible regularization strategy, support of asynchronous updates (to handle communication delays), and multiple updates per iteration (to leverage resource-rich clients) in hierarchical FL. 

In addition to these challenges, non-smooth and non-convex formulations have gained increasing importance in FL due to real-world complexities such as noise, outliers, and the need for personalization \cite{yu2024clustered,mirzaeifard2024decentralized,mirzaeifard2022robust,mirzaeifard2022dynamic}. Non-smooth objectives offer robustness against outliers and heavy-tailed distributions, improving the reliability of the learned models. Non-smooth regularization can induce sparsity (e.g., using an $l_1$ penalty) or similarity constraints (e.g., total variation norms) to promote interpretable structures \cite{mirzaeifard2024decentralized,sarcheshmehpour2023clustered}. Meanwhile, non-convex regularizers like smoothly clipped absolute deviation (SCAD) \cite{zhang2010nearly} and minimax concave penalty (MCP) \cite{fan2001variable} mitigate the bias of convex penalties and enable personalized learning by capturing intricate data relationships across heterogeneous clients \cite{yu2024clustered}. However, state-of-the-art methods often struggle with non-convexity and non-smoothness, especially when asynchronous updates, multiple local iterations, and scalability are required \cite{azimi2025hierarchical,liu2022hierarchical}. Consequently, there is a need for a new optimization framework that efficiently manages these complexities to produce robust, interpretable, and personalized solutions for large-scale hierarchical FL environments.

This work introduces the hierarchical federated smoothing ADMM (HFSAD), a framework tailored for hierarchical FL that addresses non-convex, non-smooth optimization challenges while accommodating heterogeneous network and computational conditions. HFSAD leverages smoothing techniques selectively on consensus or personalization penalties (e.g., total variation norms), making them tractable within an ADMM-based scheme. At the same time, it retains non-smooth penalties (e.g., $l_1$, MCP, SCAD) for sparsity and inductive knowledge about parameters, thereby promoting interpretability and robust regularization. This dual approach 
balances global alignment and local customization, while enabling principled handling of outliers or heavy-tailed data distributions. HFSAD supports asynchronous updates and multiple local iterations, accommodating heterogeneous network and computational conditions. By uniting smoothing-driven linearization with ADMM’s decomposition capabilities, our method maintains scalability across distributed clusters, ensuring robustness to outliers, interpretability via sparse solutions, and convergence guarantees. 
Experimental results demonstrate that HFSAD surpasses existing approaches in accuracy, speed, and flexibility, establishing a new standard for distributed data analysis under diverse system constraints.

\noindent\textit{\textbf{Mathematical Notations}}: Scalars are denoted by lowercase letters, column vectors by bold lowercase letters, and matrices by bold uppercase letters. The transpose of a matrix $\mathbf{A}$  is indicated by $\mathbf{A}^\top$. The $j$th column of $\mathbf{A}$ is denoted by $\mathbf{a}_{j}$, and the element in the $i$th row and $j$th column of $\mathbf{A}$ is denoted by $a_{ij}$. The proximal operator of a function $f(\cdot)$, scaled by a parameter $\gamma$, is defined as 
\(
\textbf{Prox}_{f}\mathopen{}\left(w;\gamma\right)\mathclose{}= \argmin_x \mathopen{}\left \{f(x)+\frac{1}{2\gamma}\left\|x-w\right\|_2^2\right \}.
\)
Finally, $\partial f(u)$ denotes the sub-gradient of the function $f(\cdot)$ at point $u$.

\section{Problem Formulation}
Assume that the system consists of one cloud server and \(L\) edge servers, where each edge server, indexed by \(l\) with \(l = 1,2,\ldots,L\), is associated with a unique set of clients \(V_l\). Each set \(V_l\) contains \(N_l\) clients, and every client \(j\) in \(V_l\) possesses its own local dataset 
$D_{lj} = \{\mathbf{X}_{lj}, \mathbf{y}_{lj}\}.$  
Consequently, the global objective function is defined as follows:
\begin{equation}\label{eq1}
    \min_{\{\{\mathbf{w}^j_{l}\}_{j\in \mathcal{N}_l}\}_{l=1}^L} \sum_{l=1}^{L} \sum_{j\in V_l} f_{lj}(\mathbf{w}^j_{l}) + \mathcal{R}_c(\{\mathbf{w}^j_{l}\}) + \mathcal{R}_w(\{\mathbf{w}^j_{l}\}),
\end{equation}
where \(\{\{\mathbf{w}^j_{l}\}_{j\in \mathcal{N}_l}\}_{l=1}^L\) denotes the local model parameters across clients. Further, within each cluster \(V_l\), there is a cluster-level parameter \(\mathbf{w}_l\) to which  individual client parameters are aligned. The regularization function \(\mathcal{R}_c(\cdot)\) is designed to enforce global consensus or allow for personalization within a cluster, while \(\mathcal{R}_w(\cdot)\) imposes prior knowledge about the parameters (e.g., based on historical data or domain expertise).

We now extend our formulation by introducing a set of cluster-level parameters \(\{\mathbf{w}_l\}_{l=1}^{L}\) and an overall global parameter \(\mathbf{w}_0\). With these additional parameters, the objective function is rewritten as:
\begin{multline}\label{eq:global2}
    \min_{\mathbf{W}} \quad \sum_{l=1}^{L} \sum_{j\in V_l} f_{lj}(\mathbf{w}^j_{l}) + \mathcal{R}_c\Bigl(\{\mathbf{w}^j_l\}_{j=1}^{N_l}, \mathbf{w}_l\Bigr) \\[1mm]
    +\, \mathcal{R}_c\Bigl(\{\mathbf{w}_l\}_{l=1}^{L}, \mathbf{w}_0\Bigr) + \mathcal{R}_w(\mathbf{w}_0) + \mathcal{R}_w\Bigl(\{\mathbf{w}_l\}_{l=1}^{L}\Bigr).
\end{multline}
Here, \(\mathbf{W}\) denotes the collection of all parameters, including the local client parameters \(\{\mathbf{w}^j_{l}\}\), the cluster-level parameters \(\{\mathbf{w}_l\}_{l=1}^{L}\), and the overall global parameter \(\mathbf{w}_0\). In this formulation, the first regularization term enforces intra-cluster relationships (which may include personalization), while the second term captures the relationship between the cluster-level parameters and the overall global parameter.

To further refine the model, we decompose the regularization into per-cluster components by defining
\(
\mathcal{R}_c^l(\mathbf{w}_l,\mathbf{w}_0),
\)
that quantifies the discrepancy between the cluster-level parameter \(\mathbf{w}_l\) and the global parameter \(\mathbf{w}_0\) for the \(l\)th cluster. Moreover, we allow the intra-cluster regularization to be flexible enough to enable personalization for each client by denoting it as
\(
\mathcal{R}_c^{l,j}(\mathbf{w}_l^j,\mathbf{w}_l).
\)
This leads to the  equivalent formulation:
\begin{multline}\label{eq:global3}
    \min_{\mathbf{W}} \quad \sum_{l=1}^{L} \Biggl[ \sum_{j\in V_l} f_{lj}(\mathbf{w}^j_{l}) + \mathcal{R}_c^{l,j}(\mathbf{w}^j_{l},\mathbf{w}_l) \\[1mm]
    +\, \mathcal{R}_c^l(\mathbf{w}_l,\mathbf{w}_0) + \mathcal{R}_w^l(\mathbf{w}_l) \Biggr] + \mathcal{R}_w^0(\mathbf{w}_0).
\end{multline}
This formulation decouples the regularization terms, allowing separate control the intra-cluster consensus (or personalization) between local and cluster-level models (via \(\mathcal{R}_c^{l,j}\)) and the inter-cluster relationship with the global model (via \(\mathcal{R}_c^l\)), while incorporating prior knowledge through the \(\mathcal{R}_w\) terms.

Motivated by the need to efficiently handle non-smooth regularizers that promote sparsity, our approach leverages the Alternating Direction Method of Multipliers (ADMM). In our formulation, the regularization function \(\mathcal{R}_w\) is often non-smooth to induce sparsity in the parameters, which enhances interpretability and leverages prior knowledge. ADMM naturally accommodates such non-smooth terms through its proximal update steps, thereby eliminating the need for ad hoc modifications. Moreover, ADMM has been demonstrated to improve learning performance in terms of both convergence speed and accuracy.

In order to apply ADMM more efficiently, we introduce auxiliary variables \(\mathbf{Z}\) and \(\mathbf{Q}\) that mirror the structure of \(\mathbf{W}\). Specifically, we define
\(\mathbf{Z} =\{ \{\mathbf{z}^j_l\}_{j\in V_l,\, l=1}^L \cup \{\mathbf{z}_l\}_{l=1}^L\}, \hspace{2mm}
\mathbf{Q} = \{\{\mathbf{q}^j_l\}_{j\in V_l,\, l=1}^L \cup \{\mathbf{q}_l\}_{l=1}^L\},
\)
where the variables \(\mathbf{z}^j_l\) and \(\mathbf{q}^j_l\) are copies of the local parameters \(\mathbf{w}^j_l\) and the cluster-level parameters \(\mathbf{w}_l\), respectively, and \(\mathbf{z}_l\) and \(\mathbf{q}_l\) are copies of \(\mathbf{w}_l\) and the global parameter \(\mathbf{w}_0\). With these auxiliary variables, our ADMM-based formulation is modified as follows:
\begin{multline}\label{eq:global4}
    \min_{\mathbf{W},\mathbf{Z},\mathbf{Q}} \quad \sum_{l=1}^{L} \Biggl[ \sum_{j\in V_l} f_{lj}(\mathbf{w}^j_{l}) + \mathcal{R}_c^{l,j}(\mathbf{z}^j_{l},\mathbf{q}^j_l) \\[1mm]
    +\, \mathcal{R}_c^l(\mathbf{z}_l,\mathbf{q}_l) + \mathcal{R}_w^l(\mathbf{w}_l) \Biggr] + \mathcal{R}_w^0(\mathbf{w}_0)\\[1mm]
    \text{subject to:} \quad \mathbf{z}^j_{l}=\mathbf{w}^j_{l},\quad \mathbf{q}^j_l=\mathbf{w}_l,\quad \forall j \in [N_l],\, \forall l \in [L],\\[1mm]
    \mathbf{z}_{l}=\mathbf{w}_{l},\quad \mathbf{q}_l=\mathbf{w}_0,\quad \forall l \in [L].
\end{multline}

However, in non-convex settings ADMM generally requires that the variables updated in its second block be smooth in order to ensure convergence \cite{wang2019global, hong2015convergence, yashtini2020convergence, themelis2020douglas, mirzaeifard2022robust}, as well as to support asynchronous and multiple updates \cite{miao2024privacy}. To address this issue, we iteratively apply smoothing techniques to those non-smooth components that are updated in the second block and that could potentially impede the update process. This reformulation preserves the sparsity-inducing properties of \(\mathcal{R}_w\) while ensuring that the ADMM updates remain stable and convergent. In the next section, we provide further details.
\section{Hierarchical Federated Smoothing ADMM}
To address the challenges in optimizing non-smooth functions within the ADMM framework, we employ smoothing techniques to approximate a  function \( g \) by a family of smooth upper-bound functions \( \tilde{g} \) \cite{mirzaeifard2023smoothing}. These approximations are  beneficial in locally Lipschitz settings, ensuring continuity, differentiability, and well-behaved gradients—features that are critical for efficient optimization. Importantly, as the smoothing parameter \(\mu\) tends to zero, \( \tilde{g}(\mathbf{x};\mu) \) converges to the original function \( g(\mathbf{x}) \), thereby preserving its essential properties \cite{chen2012smoothing}.

Specifically, we approximate the intra-cluster regularization functions \(\mathcal{R}_c^{l,j}(\mathbf{z}^j_{l},\mathbf{q}^j_l)\) and the inter-cluster regularization functions \(\mathcal{R}_c^l(\mathbf{z}_l,\mathbf{q}_l)\) by their smooth counterparts \(\tilde{\mathcal{R}}_c^{l,j}(\cdot;\mu)\) and \(\tilde{\mathcal{R}}_c^l(\cdot;\mu)\), respectively. Moreover, in cases where the loss functions \(f_{lj}(\cdot)\) are not proximal-friendly, we employ smooth approximations \(\tilde{f}_{lj}(\cdot;\mu)\) to enable a linearized update step. We emphasize that these smooth approximations serve as upper bounds to the original functions, converging to the true function as \(\mu\) decreases.

With these definitions, the approximate augmented Lagrangian for our problem can be written as
\begin{multline}
\bar{\mathcal{L}}_{\sigma,\mu}(\mathbf{W},\mathbf{Z},\mathbf{Q},\boldsymbol{\Lambda},\boldsymbol{\Gamma}) =\;  \sum_{l=1}^{L} \Biggl\{ \sum_{j\in V_l} {f}_{lj}(\mathbf{w}^j_l) + \tilde{\mathcal{R}}_c^{l,j}(\mathbf{z}^j_l,\mathbf{q}^j_l;\mu_l^j) \\
 +\, \tilde{\mathcal{R}}_c^l(\mathbf{z}_l,\mathbf{q}_l;\mu_l) + \mathcal{R}_w^l(\mathbf{w}_l) \Biggr\} + \mathcal{R}_w^0(\mathbf{w}_0) 
 + \sum_{l=1}^{L} \Biggl\{ \sum_{j\in V_l} \Biggl[\\ 
  \langle \Lambda^j_l,\, \mathbf{w}^j_l - \mathbf{z}^j_l \rangle + \frac{\sigma_l^j}{2} \|\mathbf{w}^j_l - \mathbf{z}^j_l\|^2+\langle \Gamma^j_l,\, \mathbf{w}_l - \mathbf{q}^j_l \rangle + \frac{\sigma_l^j}{2} \|\mathbf{w}_l - \mathbf{q}^j_l\|^2 \Bigr] +\\
  \langle \Lambda_l,\, \mathbf{w}_l - \mathbf{z}_l \rangle + \frac{\sigma_l}{2} \|\mathbf{w}_l - \mathbf{z}_l\|^2 +\langle \Gamma_l,\, \mathbf{w}_0 - \mathbf{q}_l \rangle + \frac{\sigma_l}{2} \|\mathbf{w}_0 - \mathbf{q}_l\|^2 \Biggr\}
\end{multline}
Here, \(\boldsymbol{\Lambda} = \{\Lambda^j_l,\, \Lambda_l\}\) and \(\boldsymbol{\Gamma} = \{\Gamma^j_l,\, \Gamma_l\}\) are the collections of dual variables
and \(\{\sigma_l^j, \sigma_l\}\) are the penalty parameters. In our algorithm, each client \(j \in V_l\) and each cluster head \(l\) maintains its own update counter—denoted by \(k_{l}^j\) for clients and \(k_{l}\) for cluster heads—that reflects the number of updates it has performed. Nodes are allowed to update asynchronously and can execute multiple updates within each global iteration \(k_{0}\). This design enables each node to progress at its own pace based on available computational resources and the characteristics of its local data, a flexibility that is essential for accommodating the inherent heterogeneity in federated learning environments. 

To progressively adjust the approximations, the parameters \( \mu_l^j, \sigma_l^j \) and \( \mu_l,\sigma_l \) are updated at each local iteration \( k_l^j \) and \( k_l^j \) respectively as following, where \( c,d,\alpha,  \beta > 0 \):
\begin{equation}\label{eq:up:sm}
\sigma_l^j{}^{(k_l^j)} = c \sqrt{k_l^j}, 
\mu_l^j{}^{(k_l^j)} = \frac{\alpha}{\sqrt{k_l^j}}, 
\sigma_l^{(k_l)} = d \sqrt{k_l}, 
\mu_l^{(k_l)} = \frac{\beta}{\sqrt{k_l}}
\end{equation}

By assuming that each function $f_{lj}(\cdot)$ is proximable and weakly convex, we update each \( \mathbf{w}_l^j \) as:
\begin{equation}\label{up:wp}
\mathbf{w}_{l}^J {}^{\left(k_l^j\right)}= \textbf{Prox}_{f_{lj}}(\boldsymbol{\iota}_l;\frac{1}{\sigma_{l}^j{}^{(k_l^j)}})
\end{equation}
where $\boldsymbol{\iota}_l=\mathbf{z}_l^j{}^{(k_l^j-1)}-\frac{\boldsymbol{\Lambda}^j_l {}^{(k_l^j-1)}}{\sigma_{l}^j {}^{k_l^j}}$. If a closed-form solution for the proximal operator of any \( f_{lj}(\cdot) \) is not available, we can employ an upper-bound smooth approximation, as suggested in \cite{mirzaeifard2022admm}, and linearize the update step.

Subsequently, in the next step, we update each \(\mathbf{z}_l^j\) and \(\mathbf{q}_l^j\) in a single step as:
\begin{multline}
  \begin{bmatrix}
 \mathbf{z}_{l}^j {}^{(k_l^j)}\\
    \mathbf{q}_{l}^j {}^{(k_l^j)}
\end{bmatrix} =\argmin_{\mathbf{z}_{l}^j,\mathbf{q}_{l}^j}  \tilde{\mathcal{R}}_c^{l,j}(\mathbf{z}^j_l,\mathbf{q}^j_l;\mu_l^j{}^{(k_l^j)})+\frac{\sigma_l^j{}^{(k_l^j)}}{2}\times\Bigg( \Bigg\| \mathbf{z}_{l}^j-\\ 
  \mathbf{w}_{l}^j {}^{(k_l^j)}-\frac{\boldsymbol{\Lambda}_{l}^{j}{}^{(k_l^j-1)}}{\sigma_l^j{}^{(k_l^j)}}\Bigg\|^2_2+ \left\|\mathbf{q}_{l}^j-\mathbf{w}_{l}{}^{(k_l)}-\frac{\boldsymbol{\Gamma}_{l}^{j}{}^{(k_l^j-1)}}{\sigma_l^j{}^{(k_l^j)}}\right\|^2_2\Bigg)  
\end{multline}
Following the simplification provided in \cite{hallac2017network}, we obtain:
\begin{dmath}\label{up:g}
     \begin{bmatrix}
 \mathbf{z}_{l}^j {}^{(k_l^j)}\\
    \mathbf{q}_{l}^j {}^{(k_l^j)}
\end{bmatrix}= 
\phantom{=}\frac{1}{2}\begin{bmatrix}
  \mathbf{w}_{l}^j {}^{(k_l^j)}+\frac{\boldsymbol{\Lambda}_{l}^{j}{}^{(k_l^j-1)}}{\sigma_l^j{}^{(k_l^j)}}+\mathbf{w}_{l}{}^{(k_l)}+\frac{\boldsymbol{\Gamma}_{l}^{j}{}^{(k_l^j-1)}}{\sigma_l^j{}^{(k_l^j)}}\\
\mathbf{w}_{l}^j {}^{(k_l^j)}+\frac{\boldsymbol{\Lambda}_{l}^{j}{}^{(k_l^j-1)}}{\sigma_l^j{}^{(k_l^j)}}+\mathbf{w}_{l}{}^{(k_l)}+\frac{\boldsymbol{\Gamma}_{l}^{j}{}^{(k_l^j-1)}}{\sigma_l^j{}^{(k_l^j)}}
\end{bmatrix}
+\frac{1}{2}\begin{bmatrix}
 -\mathbf{e}\\
\mathbf{e}
\end{bmatrix}
\end{dmath}
where
$
\mathbf{e} = \textbf{Prox}_{\tilde{\mathcal{R}}_c^{l,j}(\cdot,\cdot;\mu_l^j{}^{(k_l^j)})}\bigg(-\mathbf{w}_{l}^j {}^{(k_l^j)}-\frac{\boldsymbol{\Lambda}_{l}^{j}{}^{(k_l^j-1)}}{\sigma_l^j{}^{(k_l^j)}}+\mathbf{w}_{l}{}^{(k_l)}+\frac{\boldsymbol{\Gamma}_{l}^{j}{}^{(k_l^j-1)}}{\sigma_l^j{}^{(k_l^j)}};
\frac{2}{\sigma_l^j{}^{(k_l^j)}}
\bigg)$. When \(\mathcal{R}_c^{l,j}(\cdot)\) is defined using \(\ell_1\), MCP, or SCAD penalties, the proximal operator \(\textbf{Prox}_{\tilde{\mathcal{R}}_c^{l,j}(\cdot,\cdot;\mu_l^j)}\) is decomposable and can be efficiently computed.

Next, we update each \(\boldsymbol{\Lambda}_{l}^{j}\) and \(\boldsymbol{\Gamma}_{l}^{j}\) as follows:
\begin{equation}\label{eq:up:gl}
\boldsymbol{\Lambda}_{l}^{j}{}^{(k_l^j)}=\boldsymbol{\Lambda}_{l}^{j}{}^{(k_l^j-1)}+\sigma_l^j {}^{(k_l^j)}\left(\mathbf{w}_{l}^j {}^{(k_l^j)}-\mathbf{z}_{l}^j{}^{(k_l^j)} 
  \right)
\end{equation} 
\begin{equation} \label{eq:up:gj}
\boldsymbol{\Gamma}_{l}^{j}{}^{(k_l^j)}=\boldsymbol{\Gamma}_{l}^{j}{}^{(k_l^j-1)}+\sigma_l^j {}^{(k_l^j)}\left(\mathbf{w}_{l} {}^{(k_l)}-\mathbf{q}_{l}^j{}^{(k_l^j)} 
  \right)
 \end{equation}
 Next, we transmit the updated values \( k_l^j \), \( \mathbf{q}_{l}^j{}^{(k_l^j)} \), and \( \boldsymbol{\Gamma}_{l}^{j}{}^{(k_l^j)} \) to the cluster head.

 At the cluster head \( l \), we update \( \mathbf{w}_l \) as: 
 \begin{equation}\label{ws}
\mathbf{w}_l^{(k_l)} = \text{prox}_{\mathcal{R}_w^l}\left(\bar{\mathbf{r}}_l,\upsilon_l\right),
 \end{equation}
where $\upsilon_l=\frac{1}{\sigma_l^{k_l}+\sum_{j\in V_l}\sigma_l^j{}^{(k_l^j+1)}}$ and $\bar{\mathbf{r}}_l =\upsilon_l  \bigg( \sigma_l{}^{(k_l)}\mathbf{z}_l^{(k_l-1)} - \Lambda_l{}^{(k_l-1)} +   \sum_{j\in V_l} ( \sigma_l^j{}^{(k_l^j+1)}\mathbf{q}^j_l{}^{(k_l^j)}-\Gamma^j_l{}^{(k_l^j)}) \bigg).$ The cluster head sends its variable to each client.

Then, we update each \(\mathbf{z}_l\) and \(\mathbf{q}_l\) in a parallel as: 
\begin{multline}\label{eq13}
  \begin{bmatrix}
 \mathbf{z}_{l}{}^{(k_l)}\\
    \mathbf{q}_{l} {}^{(k_l^l)}
\end{bmatrix} =\argmin_{\mathbf{z}_{l},\mathbf{q}_{l}}  \tilde{\mathcal{R}}_c^{l}(\mathbf{z}_l,\mathbf{q}_l;\mu_l{}^{(k_l)})+\frac{\sigma_l{}^{(k_l)}}{2}\times\Bigg( \Bigg\| \mathbf{z}_{l}-\\ 
  \mathbf{w}_{l} {}^{(k_l)}-\frac{\boldsymbol{\Lambda}_{l}^{j}{}^{(k_l-1)}}{\sigma_l{}^{(k_l)}}\Bigg\|^2_2+ \left\|\mathbf{q}_{l}-\mathbf{w}_{0}{}^{(k_0)}-\frac{\boldsymbol{\Gamma}_{l}{}^{(k_l-1)}}{\sigma_l{}^{(k_l)}}\right\|^2_2\Bigg),  
\end{multline}
which can be driven in closed form solution similar to \eqref{up:g}.

The updates for dual variables $\boldsymbol{\Lambda}_l$ and $\boldsymbol{\Gamma}_{l}$ are given by
\begin{equation}\label{eq:up:gl:s}
\boldsymbol{\Lambda}_{l}^{(k_l)}=\boldsymbol{\Lambda}^{(k_l-1)}_{l}+\sigma_l^{(k_l)}\left(\mathbf{w}^{(k_l)}_l-\mathbf{z}^{(k_l)}_{l}\right)
\end{equation} 
\begin{equation} \label{eq:up:gj:s}
\boldsymbol{\Gamma}_{l}^{(k_l)}=\boldsymbol{\Gamma}^{(k_l-1)}_{l}+\sigma_l^{(k_l)}\left(\mathbf{w}_0^{(k_l)}-\mathbf{q}^{(k_l)}_l\right)
 \end{equation}
Accordingly, we send the updated values \( k_l \), \( \mathbf{q}_{l}{}^{(k_l)} \), and \( \boldsymbol{\Gamma}_{l}{}^{(k_l)} \) to server.
 
The update step for \( \mathbf{w}_0 \) is performed globally, based on aggregated information from all cluster head data after updating their respective variables. The update rule for \( \mathbf{w}_0 \) is given by: 
\begin{equation}\label{Eq.8}
   \mathbf{w}^{(k_0)} = 
\text{Prox}_{R_{w}^0(\cdot)}\left(\xi  \sum_{l=1}^L(\sigma_l^{(k_l+1)}\mathbf{q}_l^{(k_l)}+\boldsymbol{\Gamma}^{(k_l)}_l);\xi\right),
\end{equation}
where $\xi=\frac{1}{\sum_{l=1}^L\sigma_l^{(k_l+1)}}$. The server sends its variable to each cluster head accordingly. 

\begin{algorithm}[th]
\caption{Hierarchal Federated Smoothing ADMM}
\label{alg:1}
\SetAlgoLined
\medskip
\textbf{Initialization:}\\
Initialize all variables: local parameters \(\{\mathbf{w}_l^j\}\), cluster-level parameters \(\{\mathbf{w}_l\}\), global parameter \(\mathbf{w}_0\); auxiliary variables \(\mathbf{Z}, \mathbf{Q}\); and dual variables \(\boldsymbol{\Lambda}, \boldsymbol{\Gamma}\).

\For{each global iteration \(k_0 = 1,\ldots, K_z\)}{
\medskip
\For{\(k_m = 1, \ldots, K_M\)}{

\For{each cluster \(l = 1, \ldots, L\) in parallel}{
  \For{each client \(j \in V_l\) (asynchronously)}{
    \tcp{Update local smoothing parameters}
    Update \(\sigma_l^j\) and \(\mu_l^j\) according to \eqref{eq:up:sm}\;
    
    \tcp{Update local primal variable}
    Update \(\mathbf{w}_l^j\) using \eqref{up:wp} (or linearize it from its smooth approximation if necessary)\;
    
    \tcp{Update local auxiliary variables}
    Update \(\mathbf{z}_l^j\) and \(\mathbf{q}_l^j\) as in \eqref{up:g}\;
    
    \tcp{Update local dual variables}
    Update \(\boldsymbol{\Lambda}_l^j\) and \(\boldsymbol{\Gamma}_l^j\) according to \eqref{eq:up:gl} and \eqref{eq:up:gj}\;
    
    \tcp{Communicate with cluster head}
    Send updated \(\mathbf{q}_l^j\) and \(\boldsymbol{\Gamma}_l^j\) (along with the update counter \(k_l^j\)) to the cluster head.
  }}
  \For{each cluster \(l = 1, \ldots, L\) (asynchronously)}{
  \tcp{Update cluster-level smoothing parameters}
  Update \(\sigma_l\) and \(\mu_l\) using \eqref{eq:up:sm}\;
  
  \tcp{Update cluster-level variable}
  Update \(\mathbf{w}_l\) as \eqref{ws} and send it to clients.\;
  
  \tcp{Update cluster-level auxiliary variables}
  Update \(\mathbf{z}_l\) as \eqref{eq13} (similar to \eqref{up:g} for the cluster level)\;
  
  \tcp{Update cluster-level dual variables}
  Update \(\boldsymbol{\Lambda}_l\) and \(\boldsymbol{\Gamma}_l\) as per \eqref{eq:up:gl:s} and \eqref{eq:up:gj:s}\;
  
  \tcp{Communicate with central server}
  Send updated \(\mathbf{q}_l\) and \(\boldsymbol{\Gamma}_l\) (and the cluster counter \(k_l\)) to the central server.
}
}
\medskip
\textbf{Central Server Updates:}\\
    Receive updated variables from all cluster heads\;
    
    \tcp{Update global variable}
    Update \(\mathbf{w}_0\) according to \eqref{Eq.8}\;
    
    \tcp{Broadcast global variable}
    Send the updated \(\mathbf{w}_0\) to all cluster heads.}
\end{algorithm}

We emphasize that the asynchronicity and multiple updates inherent in our framework do not compromise convergence. This robustness stems from the fact that both the client-level and cluster-head dual variables are updated via an ascent step only after their corresponding primal variables (e.g., \(\mathbf{z}_l\) and \(\mathbf{q}_l\)) have been updated. Such an update order minimizes the disruptive effect of the dual updates and the associated penalty parameters, introducing only minor, summable noise due to smoothing. Moreover, although clients and cluster heads update asynchronously and independently, we enforce periodic synchronization. In particular, during a global update of \(\mathbf{w}_0\), local updates (for \(\mathbf{z}_l\), \(\mathbf{q}_l\), \(\boldsymbol{\Lambda}_l\), and \(\boldsymbol{\Gamma}_l\)) are temporarily paused, and similarly, the global update step is suspended while local nodes perform their updates. We further assume the existence of an integer \(K_a\) such that, within every \(K_a\) global iteration, every client and cluster head updates its variables at least once. 
The following theorem is provided in case consensus is needed for hierarchical learning.
\begin{theorem}\label{theorem1}
Assume that the each function $f_{lj}(\cdot)$, $R_w^l(\cdot), \forall l \in [L]\cup0$ has  bounded gradients indicated by $\nu_f$ and $\nu_r$ respectively, parameters $\sigma_l^j, 
\mu_l^j, 
\sigma_l, 
\mu_l$ are updated according to \eqref{eq:up:sm},  $\alpha c \geq \sqrt{20}\nu_f$ and $\beta d  \geq \sqrt{20}(\omega)$ where $\omega=\max_l|N_l|\nu_f+\nu_r$. 
By choosing $\mathcal{R}_c^{l,j}(\cdot)$ and $\mathcal{R}_c^{l}(\cdot)$ are total variation norm $\nu_f\|\cdot\|_1$ and $\omega\|\cdot\|_1$, Algorithm \ref{alg:1} reach consensus and converges to a stationary point satisfying the following optimality conditions:
\begin{subequations}
\begin{align}
& \mathbf{0} \in \sum_{l\in [L]}\sum_{j\in N_l} f_{lj}(\mathbf{w}_l^j {}^{*})+R_w^l(\mathbf{w}_l^{*})+R_w^0(\mathbf{w}^{*})\\ 
&\mathbf{w}_{l}{}^{*} = \mathbf{w}_l^j{}^{*}, \quad \forall j \in N_l \quad \forall l \in [L]\\
&\mathbf{w}_{0}{}^{*} = \mathbf{w}_l{}^{*}. \quad  \forall l \in [L]
\end{align}
\end{subequations}
\end{theorem}

\section{Simulation Results}
We consider the restricted-domain SCAD-penalized robust phase retrieval, where each client $j \in V_l$ optimizes:
\(f_{lj}(\mathbf{w}) = |y_{lj} - |\langle \mathbf{x}_{lj},\mathbf{w}\rangle|^2| + \mathcal{I}(\mathbf{w}),\)
with observations $\mathbf{x}_{lj} \in \mathbb{R}^M$, intensity $y_{lj} \in \mathbb{R}$, and the indicator function $\mathcal{I}(\mathbf{w})$ equals $0$ if $|w_m| \leq 5$ for all $m \in [M]$, and $+\infty$ otherwise.
The SCAD penalty is $\eta_l P_{\lambda,\gamma}(\cdot)$ with $\lambda=0.1$, $\gamma=2.4$, and weighting $\eta_l = |N_l| - \frac{1}{L}$ for $R_w^l(\cdot)$ and $\eta_l=1$ for $R_w^0(\cdot)$. Consensus is enforced via total variation regularization:
\(
\mathcal{R}_c^{l,j}(\cdot) = \omega\|\cdot\|_1, \quad \mathcal{R}_c^{l}(\cdot) = \omega_0\|\cdot\|_1,
\)
where $\omega$ is set by $5 \max_{l,j} \|\mathbf{x}_{lj}\|$, and $\omega_l = \max_l |N_l| \omega + \eta_l \lambda \gamma$. We benchmark against the centralized sub-gradient method in \cite{davis2018subgradient}, evaluating performance with relative error:
\(
\frac{\|\hat{\mathbf{x}}-\mathbf{x}\|_2^2}{\|\mathbf{x}\|_2^2},
\)
averaged over $100$ trials. Simulation parameters include: $c=\omega$, $d=\frac{\omega_0}{25}$, $\alpha=\sqrt{20}$, and $\beta=25\sqrt{20}$. Measurements are generated as:
\(
y_{lj} = |\langle \mathbf{x}_{lj}^{\top},\mathbf{w}\rangle|^2+\epsilon_{lj}, \quad \mathbf{x}_{ij}\sim \mathcal{N}(0,\mathbf{I})\odot \mathbf{B}^{l}_h,\quad \mathbf{w}\sim \mathcal{N}(0,\mathbf{I})\odot \mathbf{B}_s,
\)
where $\odot$ denotes the Hadamard product, diagonal matrices $\mathbf{B}_h^{l}, \mathbf{B}_s \in \mathbb{R}^{M\times M}$ have sparsity ratios $p=0.8$, $s=0.3$, and noise $\epsilon_{lj}$ is drawn i.i.d. from a mixture exponential distribution with parameters $c_1=0.9$, $c_2=0.1$, and $\lambda_2=\frac{\lambda_1}{10}$. The SNR ($\gamma=-20$dB) determines $\lambda_1$ via:
\(
\lambda_1=\sqrt{\frac{(\sum_{l=1}^L N_l) \times 21.8 \times 10^{\frac{\gamma}{10}}}{\sum_{l=1}^{L}\sum_{j=1}^{N_l}|\mathbf{x}_{lj}^{\top} \mathbf{w}|^2}}.
\)
In simulations, we set $(L,N,M)=(5,50,25)$, with $N_l=N$.

\begin{figure*}[t!]
  \centering
  \begin{minipage}{0.32\textwidth}
    \centering
\includegraphics[width=\textwidth]{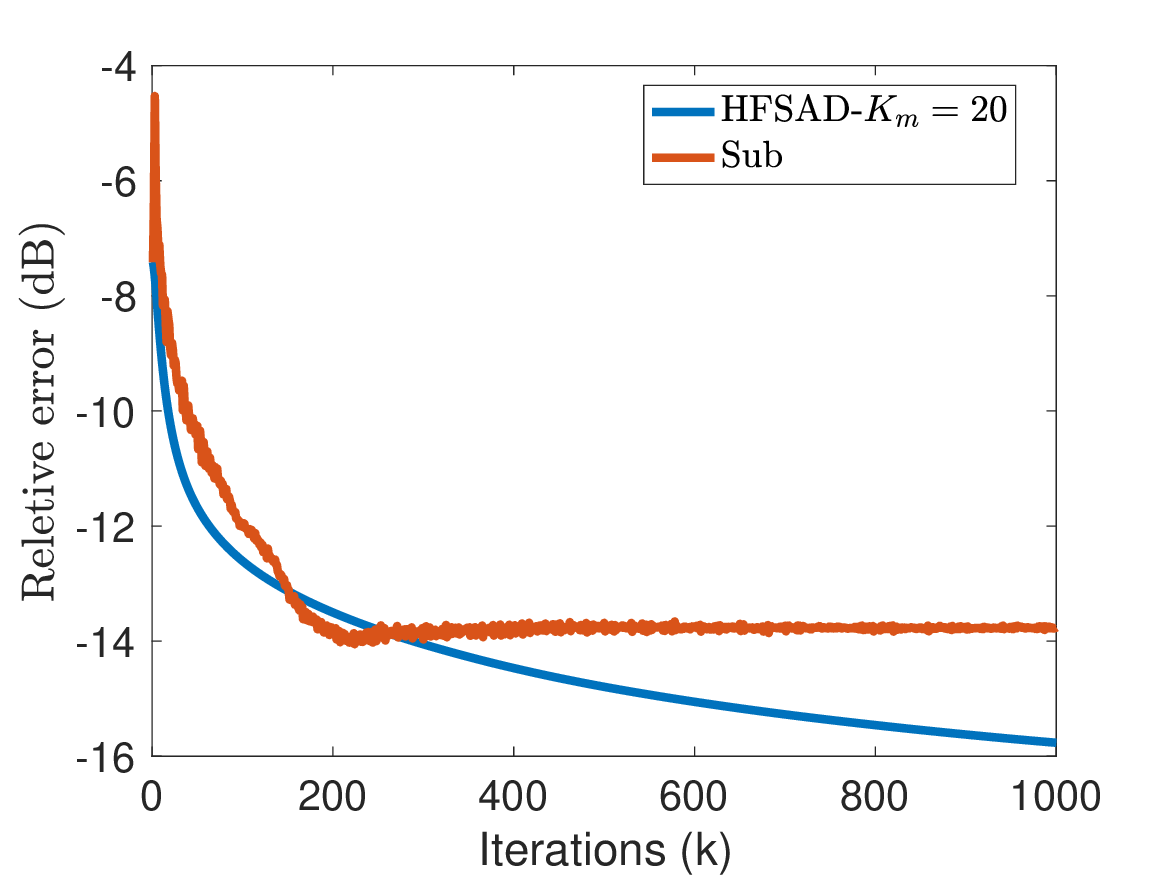}
    \caption{Relative error comparison of HFSAD and Sub} \label{fig:t3}
    \label{fig1:1}
  \end{minipage}\hfill
  \begin{minipage}{0.32\textwidth}
    \centering
    \includegraphics[width=\textwidth]{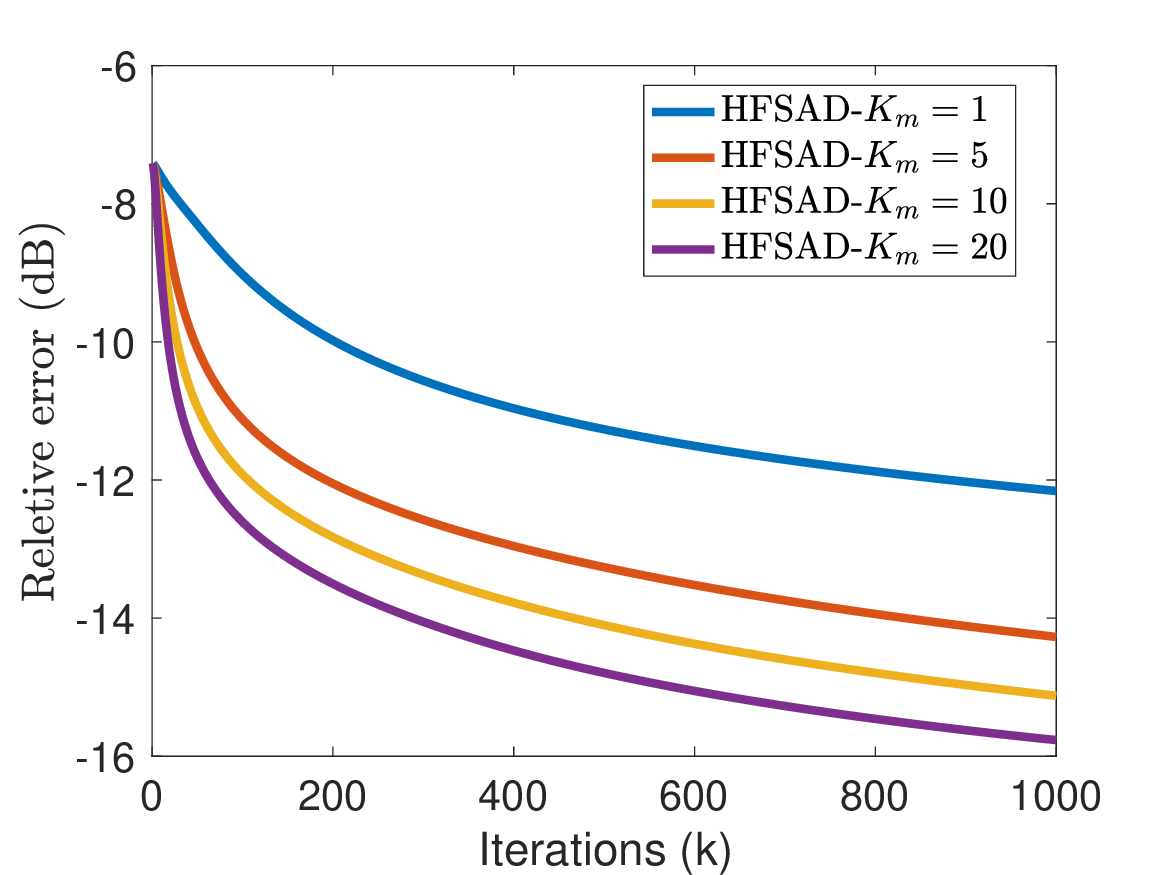}
    \caption{Relative error comparison across different update steps} \label{fig:t5}
    \label{fig1:2}
    \label{fig3}
  \end{minipage}\hfill
  \begin{minipage}{0.32\textwidth}
    \centering
\includegraphics[width=\textwidth]{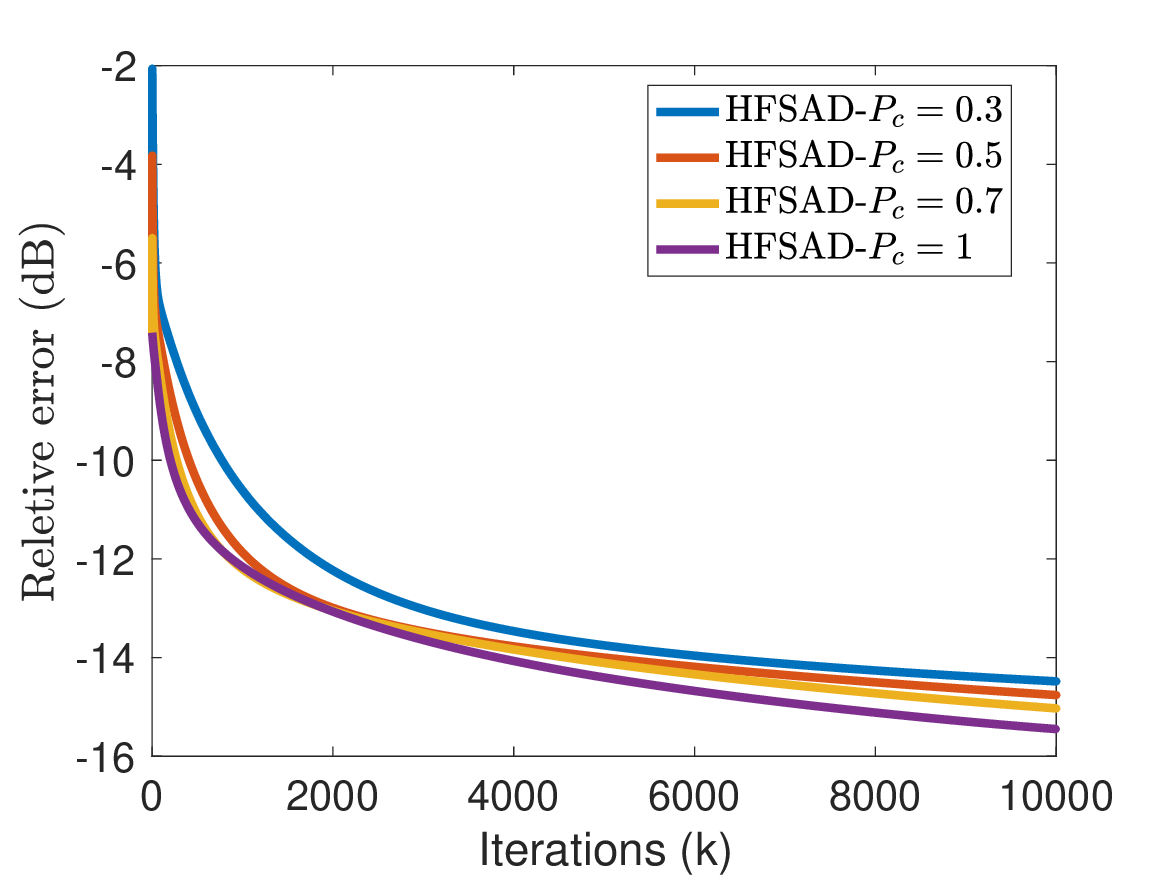}
    \caption{Relative error comparison across different collaboration probability} \label{fig:t7}
    \label{fig1:3}
  \end{minipage}
\end{figure*}

In scenario one (Fig. \ref{fig1:1}), HFSAD with $K_m=10$ local updates per iteration shows superior convergence and lower relative error compared to Sub. Scenario two (Fig. \ref{fig1:2}) compares varying local update steps ($K_m = 1,5,10,20$), demonstrating fastest convergence with $K_m=20$, while fewer steps yield slower but adequate convergence. Scenario three (Fig. \ref{fig1:3}) examines HFSAD performance under asynchronicity with probabilities $p_c = 0.3,0.5,0.7,1$ when $K_m=1$, showing comparable and acceptable performance results between different level of collaboration probabilities.

\section{Conclusion}
This paper introduced the Hierarchical Federated Smoothing ADMM (HFSAD), a novel federated learning framework designed to address non-convex and non-smooth optimization challenges in hierarchical federated learning scenarios. The proposed method integrates smoothing techniques with ADMM, effectively accommodating asynchronous updates and supporting multiple local updates per iteration, which makes it well-suited for heterogeneous network and computational environments. Our experiments on SCAD-penalized robust phase retrieval problems demonstrated that HFSAD achieves improved convergence behavior and higher accuracy compared to traditional centralized methods. These results highlight HFSAD's potential to efficiently handle complex optimization landscapes while offering robustness and flexibility required for real-world federated learning applications.

\bibliographystyle{IEEEtran}
\bibliography{strings,refs}

\begin{thebibliography}{10}
\providecommand{\url}[1]{#1}
\csname url@samestyle\endcsname
\providecommand{\newblock}{\relax}
\providecommand{\bibinfo}[2]{#2}
\providecommand{\BIBentrySTDinterwordspacing}{\spaceskip=0pt\relax}
\providecommand{\BIBentryALTinterwordstretchfactor}{4}
\providecommand{\BIBentryALTinterwordspacing}{\spaceskip=\fontdimen2\font plus
\BIBentryALTinterwordstretchfactor\fontdimen3\font minus \fontdimen4\font\relax}
\providecommand{\BIBforeignlanguage}[2]{{%
\expandafter\ifx\csname l@#1\endcsname\relax
\typeout{** WARNING: IEEEtran.bst: No hyphenation pattern has been}%
\typeout{** loaded for the language `#1'. Using the pattern for}%
\typeout{** the default language instead.}%
\else
\language=\csname l@#1\endcsname
\fi
#2}}
\providecommand{\BIBdecl}{\relax}
\BIBdecl

\bibitem{chen2019quantile}
X.~Chen, W.~Liu, and Y.~Zhang, ``Quantile regression under memory constraint,'' \emph{The Annals of Statistics}, vol.~47, no.~6, pp. 3244--3273, Dec. 2019.

\bibitem{zhou2022admm}
X.~Zhou and Y.~Xiang, ``\text{ADMM}-based differential privacy learning for penalized quantile regression on distributed functional data,'' \emph{Mathematics}, vol.~10, no.~16, p. 2954, Aug. 2022.

\bibitem{yin2021comprehensive}
X.~Yin, Y.~Zhu, and J.~Hu, ``A comprehensive survey of privacy-preserving federated learning: A taxonomy, review, and future directions,'' \emph{ACM Computing Surveys}, vol.~54, no.~6, pp. 1--36, July 2021.

\bibitem{zhou2023decentralized}
X.~Zhou, W.~Liang, I.~Kevin, K.~Wang, Z.~Yan, L.~T. Yang, W.~Wei, J.~Ma, and Q.~Jin, ``Decentralized p2p federated learning for privacy-preserving and resilient mobile robotic systems,'' \emph{IEEE Wireless Communications}, vol.~30, no.~2, pp. 82--89, Apr. 2023.

\bibitem{chen2020asynchronous}
Y.~Chen, Y.~Ning, M.~Slawski, and H.~Rangwala, ``Asynchronous online federated learning for edge devices with non-iid data,'' in \emph{IEEE International Conference on Big Data}, 2020, pp. 15--24.

\bibitem{Kairouz2021Advances}
P.~Kairouz, H.~B. McMahan, B.~Avent, A.~Bellet, M.~Bennis, A.~N. Bhagoji \emph{et~al.}, ``Advances and open problems in federated learning,'' \emph{Foundations and Trends in Machine Learning}, vol.~14, no. 1--2, pp. 1--210, 2021.

\bibitem{Zhang2022Federated}
J.~Zhang, Z.~Li, B.~Li, J.~Xu, S.~Wu, S.~Ding, and C.~Wu, ``Federated learning with label distribution skew via logits calibration,'' in \emph{Proceedings of the 39th International Conference on Machine Learning (ICML)}, ser. Proceedings of Machine Learning Research, vol. 162, 2022, pp. 26\,311--26\,329.

\bibitem{Diao2020HeteroFL}
E.~Diao, J.~Ding, and V.~Tarokh, ``{HeteroFL}: Computation and communication efficient federated learning for heterogeneous clients,'' in \emph{International Conference on Learning Representations (ICLR)}, 2020.

\bibitem{Li2019FedMD}
D.~Li and J.~Wang, ``{FedMD}: Heterogeneous federated learning via model distillation,'' \emph{arXiv preprint arXiv:1910.03581}, 2019.

\bibitem{Smith2017Federated}
V.~Smith, C.~Chiang, M.~Sanjabi, and A.~Talwalkar, ``Federated multi-task learning,'' in \emph{Advances in Neural Information Processing Systems (NIPS) 30}, 2017, pp. 4424--4434.

\bibitem{Marfoq2021Federated}
O.~Marfoq, G.~Neglia, A.~Bellet, L.~Kameni, and R.~Vidal, ``Federated multi-task learning under a mixture of distributions,'' in \emph{Advances in Neural Information Processing Systems 34 (NeurIPS 2021)}, 2021.

\bibitem{Nishio2019Client}
T.~Nishio and R.~Yonetani, ``Client selection for federated learning with heterogeneous resources in mobile edge,'' in \emph{Proc. IEEE International Conference on Communications (ICC)}, 2019.

\bibitem{Lai2021Oort}
F.~Lai, X.~Zhu, H.~V. Madhyastha, and M.~Chowdhury, ``Oort: Efficient federated learning via guided participant selection,'' in \emph{15th USENIX Symposium on Operating Systems Design and Implementation (OSDI)}, 2021, pp. 19--35.

\bibitem{shin2022fedbalancer}
J.~Shin, Y.~Li, Y.~Liu, and S.-J. Lee, ``Fedbalancer: Data and pace control for efficient federated learning on heterogeneous clients,'' in \emph{Proceedings of the 20th Annual International Conference on Mobile Systems, Applications and Services}, 2022, pp. 436--449.

\bibitem{li2020federated}
T.~Li, A.~K. Sahu, A.~Talwalkar, and V.~Smith, ``Federated learning: Challenges, methods, and future directions,'' \emph{IEEE Signal Processing Magazine}, vol.~37, no.~3, pp. 50--60, May 2020.

\bibitem{yu2024clustered}
X.~Yu, Z.~Liu, W.~Wang, and Y.~Sun, ``Clustered federated learning based on nonconvex pairwise fusion,'' \emph{Information Sciences}, vol. 678, p. 120956, 2024.

\bibitem{mirzaeifard2024decentralized}
R.~Mirzaeifard, D.~Ghaderyan, and S.~Werner, ``Decentralized smoothing {ADMM} for quantile regression with non-convex sparse penalties,'' \emph{arXiv preprint arXiv:2408.01307}, 2024.

\bibitem{mirzaeifard2022robust}
R.~Mirzaeifard, N.~K. Venkategowda, and S.~Werner, ``Robust phase retrieval with non-convex penalties,'' in \emph{56th IEEE Asilomar Conference on Signals, Systems, and Computers}, 2022, pp. 1291--1295.

\bibitem{mirzaeifard2022dynamic}
R.~Mirzaeifard, V.~C. Gogineni, N.~K. Venkategowda, and S.~Werner, ``Dynamic graph topology learning with non-convex penalties,'' in \emph{30th IEEE European Signal Processing Conference}, 2022, pp. 682--686.

\bibitem{sarcheshmehpour2023clustered}
Y.~SarcheshmehPour, Y.~Tian, L.~Zhang, and A.~Jung, ``Clustered federated learning via generalized total variation minimization,'' \emph{IEEE Transactions on Signal Processing}, vol.~71, pp. 4240--4256, 2023.

\bibitem{zhang2010nearly}
C.-H. Zhang, ``Nearly unbiased variable selection under minimax concave penalty,'' \emph{The Annals of Statistics}, vol.~38, no.~2, pp. 894--942, Apr. 2010.

\bibitem{fan2001variable}
J.~Fan and R.~Li, ``Variable selection via nonconcave penalized likelihood and its oracle properties,'' \emph{Journal of the American Statistical Association}, vol.~96, no. 456, pp. 1348--1360, Dec. 2001.

\bibitem{azimi2025hierarchical}
S.~M. Azimi-Abarghouyi, N.~Bastianello, K.~H. Johansson, and V.~Fodor, ``Hierarchical federated admm,'' \emph{IEEE Networking Letters}, 2025.

\bibitem{liu2022hierarchical}
L.~Liu, J.~Zhang, S.~Song, and K.~B. Letaief, ``Hierarchical federated learning with quantization: Convergence analysis and system design,'' \emph{IEEE Transactions on Wireless Communications}, vol.~22, no.~1, pp. 2--18, 2022.

\bibitem{wang2019global}
Y.~Wang, W.~Yin, and J.~Zeng, ``Global convergence of admm in nonconvex nonsmooth optimization,'' \emph{Journal of Scientific Computing}, vol.~78, no.~1, pp. 29--63, Jan. 2019.

\bibitem{hong2015convergence}
M.~Hong, Z.-Q. Luo, and M.~Razaviyayn, ``Convergence analysis of alternating direction method of multipliers for a family of nonconvex problems,'' in \emph{IEEE International Conference on Acoustics, Speech and Signal Processing}, Apr. 2015, pp. 3836--3840.

\bibitem{yashtini2020convergence}
M.~Yashtini, ``Convergence and rate analysis of a proximal linearized {ADMM} for nonconvex nonsmooth optimization,'' \emph{Journal of Global Optimization}, vol.~84, no.~4, pp. 913--939, Dec. 2022.

\bibitem{themelis2020douglas}
A.~Themelis and P.~Patrinos, ``Douglas--{Rachford} splitting and {ADMM} for nonconvex optimization: Tight convergence results,'' \emph{SIAM Journal on Optimization}, vol.~30, no.~1, pp. 149--181, Jan. 2020.

\bibitem{miao2024privacy}
Y.~Miao, D.~Kuang, X.~Li, S.~Xu, H.~Li, K.-K.~R. Choo, and R.~H. Deng, ``Privacy-preserving asynchronous federated learning under non-iid settings,'' \emph{IEEE Transactions on Information Forensics and Security}, 2024.

\bibitem{mirzaeifard2023smoothing}
R.~Mirzaeifard, N.~K. Venkategowda, V.~C. Gogineni, and S.~Werner, ``Smoothing admm for sparse-penalized quantile regression with non-convex penalties,'' \emph{IEEE Open Journal of Signal Processing}, 2023.

\bibitem{chen2012smoothing}
X.~Chen, ``Smoothing methods for nonsmooth, nonconvex minimization,'' \emph{Mathematical Programming}, vol. 134, pp. 71--99, Aug. 2012.

\bibitem{mirzaeifard2022admm}
R.~Mirzaeifard, N.~K. Venkategowda, V.~C. Gogineni, and S.~Werner, ``\text{ADMM} for sparse-penalized quantile regression with non-convex penalties,'' in \emph{30th IEEE European Signal Processing Conference}, 2022, pp. 2046--2050.

\bibitem{hallac2017network}
D.~Hallac, Y.~Park, S.~Boyd, and J.~Leskovec, ``Network inference via the time-varying graphical lasso,'' in \emph{Proc. 23th Int. Conf. Knowl. Discovery and Data Mining}, 2017, pp. 205--213.

\bibitem{davis2018subgradient}
D.~Davis, D.~Drusvyatskiy, K.~J. MacPhee, and C.~Paquette, ``Subgradient methods for sharp weakly convex functions,'' \emph{Journal of Optimization Theory and Applications}, vol. 179, pp. 962--982, Dec. 2018.

\end{thebibliography}

\end{document}